%% file: iclr2024_conference.tex
\documentclass{article} %
\usepackage{iclr2024_conference,times}
\usepackage{subfigure}
\usepackage{subcaption}
\usepackage{enumitem}
\usepackage{graphicx}
\usepackage{multirow}

\input{math_commands.tex}

\usepackage{xcolor}
\usepackage{wrapfig}
\usepackage{hyperref}
\usepackage{url}
\usepackage{pifont}
\usepackage{booktabs} 
\usepackage{makecell}
\usepackage{caption}
\usepackage{graphicx}

\definecolor{oucrimsonred}{rgb}{0.6, 0.0, 0.0}
\definecolor{cbgreen}{RGB}{94, 201, 72}
\definecolor{cbgreen}{RGB}{0, 128, 0}
\usepackage{booktabs}
\usepackage{xcolor}
\usepackage{calc}

\newlength\WIDTHOFBAR
\title{Towards an empirical understanding of \\
MoE Design Choices}

\author{Dongyang Fan$^\star$, Bettina Messmer$^\star$, Martin Jaggi  \\
EPFL, Switzerland \\
\texttt{firstname.lastname@epfl.ch} \\
}

\iclrfinalcopy
\begin{document}

\maketitle

\section{Introduction}
The Mixture of Experts (MoEs) has been receiving unprecedented attention in the LLM era. While initially it has been proposed by \citet{jacobs1991adaptive} to encourage expert specialization when the model is under-parameterized to fit the whole data domain, the contemporary practices~\citep{fedus2022switch,shazeer2017outrageously} do not specifically seek for expert specialization aspects, instead, they use MoE as a tool to scale up model expressiveness at a reduced inference cost. A study by \cite{zoph2022st} revealed the existence of expert specialization in encoder blocks, particularly at a lexicon level.
Furthermore, the recent Mistral paper by~\cite{jiang2024mixtral} provides evidence that the router exhibits structured syntactic behavior rather than topic-level understanding. We posit that the cultivation of fine-grained expert specialization is facilitated by Token-level routing mechanisms. At the sequence level, the router is compelled to consider contextual information. A natural question arises: \emph{do experts acquire different knowledge bases when routing is performed at a sequence level?}

MoEs demonstrate exceptional performance, 
yet the specific design choices that contribute to this efficacy remain a subject of inquiry. In particular, the impact of these choices on Sequence-level routing's marginal validation performance is not well understood. \cite{jiang2024mixtral} employs a Layer-wise Token-level Top-2 routing, a prevalent architectural choice in contemporary implementations. Meanwhile, \cite{fedus2022switch} empirically substantiates the potency of Top-1 routing for Token-level routing. We wish to \emph{systematically ablate each of these design choices and discern the extent to which they contribute to model performance.}

In this study, we empirically investigate Mixture of Experts (MoE) training from scratch using GPT2 Small-scale models. While our findings may not generalize to larger Language Model scales due to computing constraints in academia, we aim to offer insights into MoEs. Our contributions include:
\setitemize{noitemsep,topsep=0pt,parsep=1.5pt,partopsep=0pt}
\begin{itemize}[leftmargin=*]
    \item Ablating common MoE design choices to quantify the marginal impact on validation performance.
    \item Demonstrating that design choice preferences might differ for Token- and Sequence-level routing.
    \item Justifying the existence of \emph{weak expert} specialization in topics when routing at a sequence level.
\end{itemize}

\section{Related Works}
\paragraph{Design Choices} \cite{shazeer2017outrageously} proposed Noisy Top-K Gating and required $K >1$ in order to receive gradients for the router network. \cite{fedus2022switch} made Top-1 routing work by multiplying Top-1 gating probability with the corresponding gating output. Both works let the token itself make a choice on which expert to go to, which usually leads to expert collapse. An auxiliary loss or special gating mechanism~\citep {lewis2021base} is needed to balance the load of the experts. While the most popular routing unit is token, some works explore sentence-level routing~\citep {pham2023taskbased, kudugunta2021exploring} and task-level routing~\citep{li2022branchtrainmerge,fan2021beyond}. Task-level routing is predominantly used in multilingual translation settings, where an extra indicator for language pairs is used. \vspace{-5pt}
\paragraph{Expert Specialization} \citet{zoph2022st} found in an encoder-decoder network, only encoder experts exhibit specialization while decoder experts lack this ability. This specialization manifests primarily in syntactic features, encompassing punctuation, conjunctions, articles, and verbs. \citet{jiang2024mixtral} corroborated the presence of structured syntactic behavior among experts but did not discern explicit patterns in expert assignments based on the topic.  \citet{openmoe2023} examined different levels of specialization and observed only context-independent specialization at a token ID level. Notably, the routing unit in these studies is at the token level.  We want to investigate the feasibility of inducing context-dependent specialization through Sequence-level routing.

\begin{table}[t!]
 \vspace{-3em}
\centering
\begin{tabular}{llll |c@{} | c@{}} 
 \toprule
\multirow{ 2}{*}{\textbf{Topk($K$)}} & \multicolumn{ 2}{c}{\textbf{Design Choices}} & {\textbf{\#Parameters}} &  \multicolumn{2}{c}{\textbf{Validation Perplexity} }\\[0.5ex]

& {\emph{Routing Unit}} & {\emph{Routing Strategy}}& \emph{(Total, Active)} & \emph{MultiWiki} \: & \emph{OpenWebText} \\
\midrule
\multirow{3}{*}{2} & Token & Layer-wise & \multirow{3}{*}{(295M, 182M)}& \textbf{\textcolor{oucrimsonred}{9.907}} & \textbf{\textcolor{oucrimsonred}{22.632}}\\ 
 & Sequence & Layer-wise & & \textbf{\textcolor{cbgreen}{10.667}}  & 24.980\\ 
  & Sequence & Global & &  11.605 & 25.689\\ 
  \midrule
  \multirow{3}{*}{1} & Token & Layer-wise & \multirow{3}{*}{(295M, 124M)} & \textbf{\textcolor{cbgreen}{10.188}} & \textbf{\textcolor{cbgreen}{23.548}}\\
    & Sequence & Layer-wise & & 11.573 & 26.742\\ 
   & Sequence & Global & & 13.632 & 30.292\\ 
  \midrule
  \midrule
  \multicolumn{ 3}{l}{Baseline - $1 \times$ FFN Width} & (124M, 124M) & 11.815 & 26.195\\
  \multicolumn{ 3}{l}{Baseline - $2 \times$ FFN Width} & (182M, 182M) & 10.831 & 24.387\\
  \multicolumn{ 3}{l}{Baseline - $4 \times$ FFN Width} & (295M, 295M) & 10.071 & 22.700 \\
  \bottomrule
 \end{tabular}
 \caption{Ablation studies of each component of MoE design choices. Number of experts ($N$) is set to 4 in all experiments. \textbf{\textcolor{oucrimsonred}{Red}} signifies surpassing the baseline model with the entire parameter count, while \textbf{\textcolor{cbgreen}{green}} indicates surpassing the baseline model with an equivalent active parameter count.}
 \label{tab: ablation}
 \vspace{-1em}
 \end{table}

\section{Experiments}
Our experiments are rooted in the GPT2-small (124M)\footnote{\url{https://github.com/karpathy/nanoGPT}} model. The MoE component is added to the feedforward layer (FFN) in \emph{every} transformer block, with an MLP router network deciding which expert(s) to route each sequence/token to. For Top-1 gating we drop the expert capacity limit in our experiments.  Our experiments were conducted using two different datasets, one is multilingual Wikipedia~\citep{wikidump} with documents in English, German, French, and Italian four languages (17B tokens), and the other is OpenWebText~\citep{Gokaslan2019OpenWeb}, which is gathered from scrapping URLs from Reddit posts and only available in English (9B tokens). We report our experimental results as the average over the last 100 iterations, to address the inability to conduct multiple runs with different seeds imposed by our computing resources limitation. See Appendix~\ref{sec:appx:tech_details} for further technical details.

\subsection{Performance Impact of Design Choices}
Note our Global Top-1 model represents the classic MoE structure from the 90s~\citep{jacobs1991adaptive}, Token-level Layer-wise Top-1 mimics the switch transformer~\citep{fedus2022switch} idea, while Token-level Layer-wise Top-2 routing represents the most popular MoE implementation nowadays~\citep{shazeer2017outrageously, lepikhin2020gshard, jiang2024mixtral}. 

The baseline methods are selected as the dense model counterparts matching the total number of parameters or the number of active parameters in the forward pass. To do so, we simply duplicate FFN blocks. Examining Table~\ref{tab: ablation}, it is evident that with the same amount of total parameters, only the combination of Token-level Layer-wise Top-2 routing surpasses the dense model baselines. Token-level Layer-wise Top-1 routing outperforms the dense model baseline matching the active parameter count by a large margin, while Sequence-level Layer-wise Top-2 routing performs comparably to its dense counterpart with an equivalent number of active parameters.
In general, we observe that the routing unit has the biggest impact on validation performance.

\subsubsection{Does Expert collapse hurt?}
\begin{figure}[t!]
\vspace{-3em}
  \centering
  \subfigure{\includegraphics[width=0.45\textwidth]{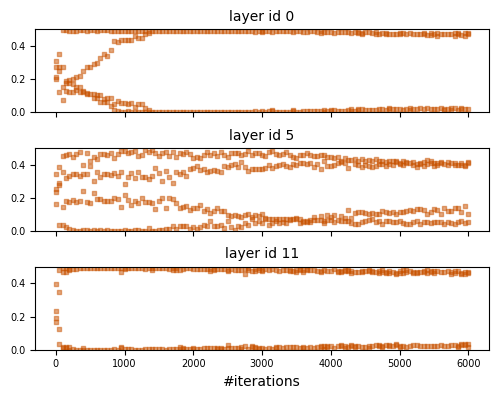}}
  \hfill
  \subfigure{\includegraphics[width=0.45\textwidth]{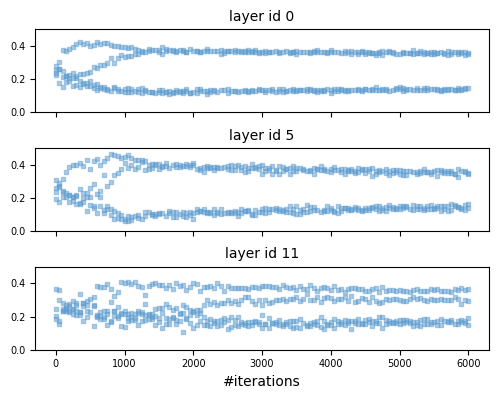}}
  \caption{The frequency that each expert is activated during training for every iteration. Left: pretraining without load balancing loss, resulting in a validation perplexity 10.674. Right: with load balancing loss ($\lambda=0.01$), resulting in a validation perplexity 10.667.}
  \label{fig:expert-collapse}
  \vspace{-1em}
\end{figure}

As previously reported in \cite{fedus2022switch, zoph2022stmoe}, expert collapse is always observed without adding an auxiliary loss. Does expert collapse indeed hurt a model's validation performance? Without any auxiliary loss, we witnessed expert collapse in early and late layers, as shown in the left panel of Figure~\ref{fig:expert-collapse}. Nonetheless, the validation perplexities with or without load balancing loss are almost identical, confirming the findings in \citet{nie2022densetosparse, yang2021m6}. This could indicate that we do not necessarily need the same amount of experts in every layer, as in \cite{pmlr-v162-du22c}. 
It is worth noting that we did not employ an expert capacity factor in any of our experiments. A more detailed analysis of expert activation is available in Appendix~\ref{append-layerwise-expert-activation}.

\subsubsection{Does more experts and more activated experts always help?}
For every sequence or token at each layer, there exist ${N \choose K}$ combinations of experts to which activations can be directed, resulting in $12^{N \choose K}$ distinct path configurations for each routed unit. An increase in the value of $N$ expands the range of routing options for each sequence, while a larger value of $K$ leads to the activation of more neurons within a layer. To investigate the effects, we replicate FFN block $N$ times, indicating $N$ experts. Our findings reveal that \emph{increased activated experts ($K$) improve performance for sequence-level routing, whereas token-level routing benefits more from a greater number of experts ($N$)}, i.e. tokens in a sequence can be routed to different experts.

\begin{wraptable}{R}{0.4\textwidth}
\vspace{-1.75em}
\begin{center}
\resizebox{1.00\linewidth}{!}{
\begin{tabular}{l | c@{}|c@{}} 
 \toprule
\multirow{2}{*}{\textbf{ ($\mathbf{ N, K }$)}} &  \multicolumn{ 2}{c}{\textbf{Validation Perplexity}}\\  
& \emph{Sequence} \:& \emph{Token}\\
 \midrule
( 2 , 2 ) & 10.648 & 10.761\\
( 4 , 2 ) & 10.667 & 9.907 \\ 
( 6 , 2 ) & 10.674 & \textbf{9.467} \\ 
( 4 , 1 ) &  11.573  & 10.188\\ 
( 4 , 3 ) & \textbf{10.107} & 9.711\\
 \bottomrule
\end{tabular}}
\caption{The impact of number of experts ($N$) and number of activated experts ($K$). Pretrained on Multilingual Wikipedia dataset.}
\end{center}
\label{tab:num_exp}
\vspace{-3em}
\end{wraptable}
Additionally, routing at a sequence level shows that the $(4,1)$ combination significantly underperforms compared to $(4,2)$, likely because it lacks dominant experts. In contrast to Token-level routing, we see the most performance gain for Top-1 routing and diminishing return for larger K's. This is consistent with the findings of ~\cite{clark2022unified, yang2021m6}. It is noteworthy that the small performance gap between Top-1 and Top-2 routing does not seem to extend to larger models as found by~\cite{yang2021m6, lewis2021base}, where the expert capacity factor is applied.

\subsection{Does expert specialization exist?}

\subsubsection{What does a router learn?}
\paragraph{Frozen routing.} To understand if the learned routers (\emph{Learned}) have learned something meaningful, we design experiments where the routing parameters were either frozen at initialization (\emph{Frozen}) or re-initialized randomly (\emph{Random}) at each iteration. It is noteworthy that frozen routing shares a conceptual similarity with hash routing. However, the routing result is embedding dependent instead of token id dependent as in \citet{roller2021hash}. That is, in our frozen routing setup, expert specialization can still be learned by letting embedding layers adapt to a frozen router head. Layer-wise frozen routing almost gives the same performance as Layer-wise learned routing, as shown in Table~\ref{tab:frozen-routing}, supporting this assumption. This finding aligns with \cite{dikkala2023benefits}. 
\begin{table}[h!]
\vspace{-3em}
\begin{center}
\begin{tabular}{l l l | c@{} |  c@{} } 
 \toprule
\multirow{2}{*}{\textbf{Dataset}} & \multirow{2}{*}{\textbf{Routing Level}}  & \multirow{2}{*}{\textbf{Routing Strategy}} &  \multicolumn{2}{c}{\textbf{Validation Perplexity}} \\
& & & \: \emph{Sequence} \: & \emph{Token}\\ 
 \midrule
Multilingual  & Global & Language  & 11.716 & -\\
&  Global & Learned &  11.605 & -\\
 & Layer-wise & Frozen &  \textbf{10.601} & \textbf{9.810}
\\ 
& Layer-wise & Random &  11.685 & 11.752\\ 
&  Layer-wise & Learned  & \textbf{10.666} &  \textbf{9.907}\\ 
\midrule
OpenWebText & Layer-wise & Frozen & \textbf{24.590} & \textbf{22.687}
\\ 
& Layer-wise & Random &  26.429 & 26.879\\ 
& Layer-wise & Learned &  \textbf{24.980} & \textbf{22.632}\\ 
 \bottomrule
\end{tabular}
\caption{Comparison of different routing strategies. \emph{Global} indicates there is only one router at the very first transformer block, and the other layers simply follow this routing output.}
\label{tab:frozen-routing}
\end{center}
\end{table}
\paragraph{Language guided routing versus learned Global routing.} In the multilingual setup, as languages are naturally different, we examine if our learned router can group languages as expected when routing at a sequence level. One interesting baseline we compared to here is hard-coded per-language routing (\emph{Language}). We simply direct sequences from different languages to different experts. To our surprise, this model does not perform as well as expected. One might intuitively anticipate that each expert, assigned a more uniform task (arguably simpler), would learn more rapidly and proficiently. Given our learned Global routing gives a roughly similar performance, we conclude that more routing path possibilities are needed, i.e. Layer-wise routing. The learned router also does not show an ability to differentiate different languages as discussed in the next Section \ref{sec:weak_expert_spec}.

\subsubsection{The existence of weak experts?}\label{sec:weak_expert_spec}

\begin{figure}
\vspace{-1em}
  \centering
  \subfigure{\includegraphics[width=0.32\textwidth]{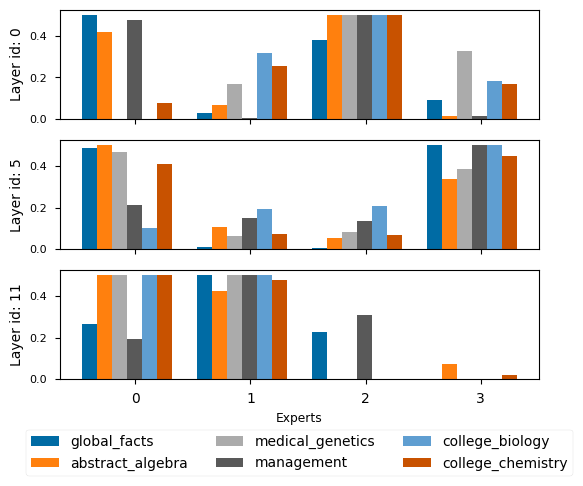}}
  \hfill
  \subfigure{\includegraphics[width=0.31\textwidth]{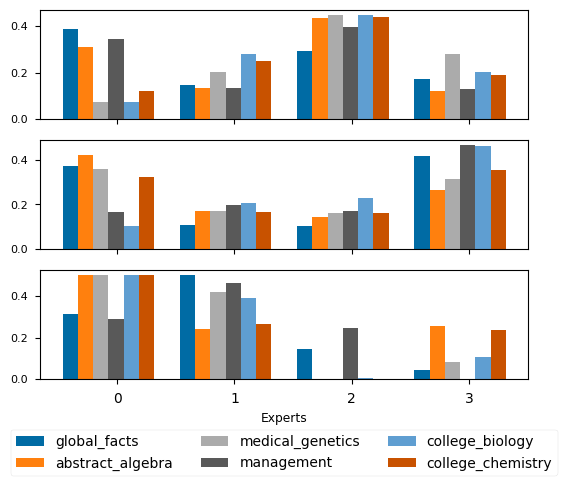}}
 \hfill
  \subfigure{\includegraphics[width=0.31\textwidth]{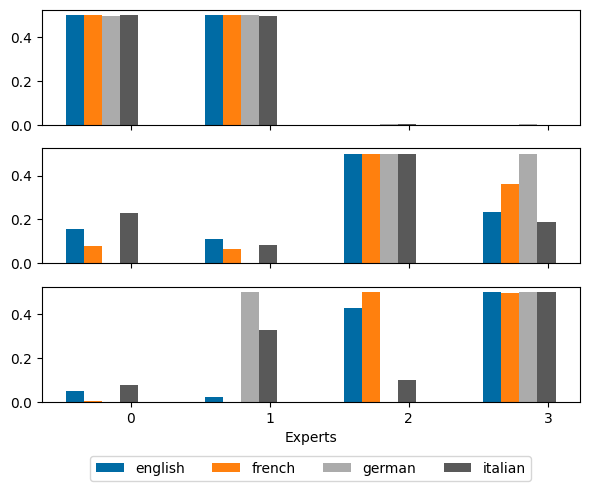}}
  
  \caption{Layer-wise expert assignment results. From left to right: (1) pretrained on OpenWebText dataset, evaluated on 6 categories of MMLU dataset; (2) pre-trained on Multilingual Wikipedia dataset, evaluated on 6 categories of MMLU dataset; (3) pre-trained on Multilingual Wikipedia dataset, evaluated on 4 languages from XNLI dataset and X-Stance Dataset.}
  \label{pic: Layer-wise expert assignment}
  \vspace{-1em}
\end{figure}

To validate the presence of expert specialization in Sequence-level routing, we assess our pre-trained model's performance on unseen tasks. We examine whether there is a discernible pattern in expert activation for distinct tasks. Six task domains from the MMLU dataset \citep{hendryckstest2021} and four languages from the XNLI dataset~\citep{conneau2018xnli} and X-stance dataset~\citep{vamvas2020xstance} were selected to determine if experts acquired domain-specific or language-specific knowledge. Irrespective of whether the model was pre-trained on English-only or multilingual datasets, the assignment of experts consistently reveals similar patterns when routing samples across different domains. For instance, there seems to be a similar routing pattern for both global facts and management. Evidently, expert assignments are language-independent, as illustrated in the rightmost panel of Figure~\ref{pic: Layer-wise expert assignment}, indicating that it might be beneficial to jointly learn concepts across similar languages. A pre-trained model with Token-level routing shows very different routing behaviors, as shown in Figure~\ref{fig: token-level-routing-specialization} in Appendix~\ref{append: expert-assignment}.

\section{Conclusion}
This study delves into the training and design choices of Mixture of Experts (MoEs), focusing on their impact on model performance and expert specialization.
Notably, the preferred design choices differ when routing at a token or sequence level: increased activated experts ($K$) improve performance for sequence-level routing, whereas token-level routing benefits more from a greater number of experts ($N$). Moreover, our findings indicate the possibility of fostering weak expert specialization related to topics through the implementation of Sequence-level routing. In practical terms, we demonstrate that achieving expert specialization during MoE training is attainable by simply initializing router weights randomly. We hope these observations can contribute to a better understanding of MoE design choices for practitioners.
\newpage

\bibliography{iclr2024_conference}
\bibliographystyle{iclr2024_conference}
\newpage
\appendix
\section{Appendix}
\subsection{Experimental Details}\label{sec:appx:tech_details}
We run our experiments based on the publicly available LoRA extension~\citep{loraNanoGPT} of the nanoGPT library~\citep{nanoGPT} on a single A100-SXM4-40GB GPU.
We use the default setup, but with dropout $0.2$, learning rate $9.6$e-4, minimum learning rate $9.6$e-5, weight decay $0.5$ and enabled biases for $6$K iterations.  Every iteration 1'048'576 tokens are passed through the network (gradient accumulation 128, batch size 8, sequence length 1024), resulting in roughly 6B tokens being seen by the model. For our default set up with 4 experts this follows the Chinchilla scaling law from~\cite{hoffmann2022training} to train on approximately 20 tokens per parameter.
Additionally, we added an optional load balancing loss with weight ($\lambda$ = 0.01) for our MoE experiments.

For both datasets we use the OpenAI GPT-2 tokenizer, which gives a slight disadvantage to the non-English languages. However, we observe that the model can still learn the non-English languages, in close language groups for next token prediction.

Assume gating network $W_g$ produces logits $h(x) = W_g · x$. Expert blocks are denoted as $E_i$s. Our Top-1 gating follows the implementation of switch transformer~\citep{fedus2022switch}. 
\begin{equation}
    p_i(x) = \frac{e^{h_i(x)}}{\sum_j e^{h_j(x)}} \qquad y = \sum_{i \in \gT} p_i(x) E_i(x)
\end{equation}

Top-2 gating implementation follows the implementation of \citet{jiang2024mixtral}, except for Sequence-Level routing we apply the softmax operation twice, as shown below:
\begin{equation}
    p_i(x) = \frac{e^{h_i(x)}}{\sum_j e^{h_j(x)}} \qquad y = \sum_{i \in \gT} \frac{e^{p_i(x)}}{\sum_{j \in \gT} e^{p_j(x)}} E_i(x)
\end{equation}
Note $\gT$ in the formulas denote the set of Top-K index/indices. 

\subsection{On Router Choices}
\begin{figure}[h!]
    \centering
    \includegraphics[width=.6\textwidth]{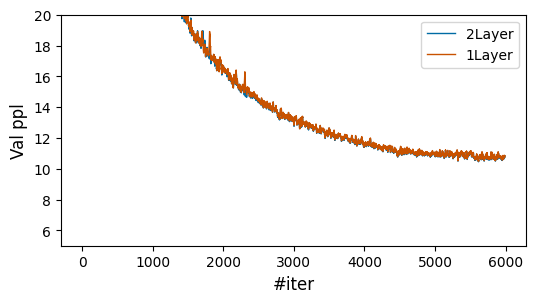}
    \caption{Validation perplexity versus training iterations.}
    \label{fig:1layer-2layer-mlp}
\end{figure}

Empirically, using one-layer MLP or two-layer MLP as the router network does not make a difference with regard to validation performances, as shown in Figure~\ref{fig:1layer-2layer-mlp}. 
\subsection{Expert Assignments}
\label{append: expert-assignment}
\begin{figure}[h!]
  \centering
   \subfigure{\includegraphics[width=0.31\textwidth]{pics/open-topic.png}}
  \subfigure{\includegraphics[width=0.3\textwidth]{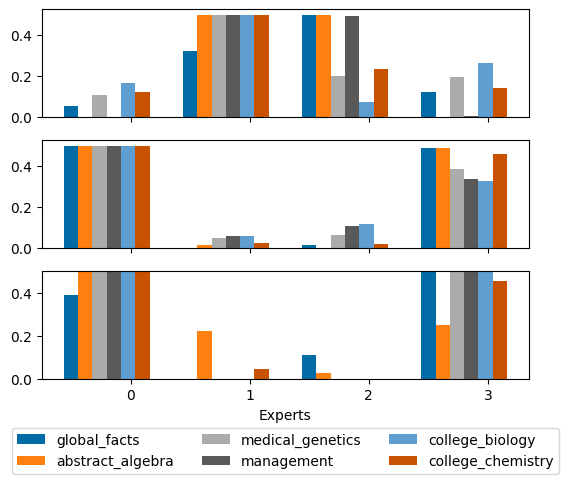}}
  \subfigure{\includegraphics[width=0.3\textwidth]{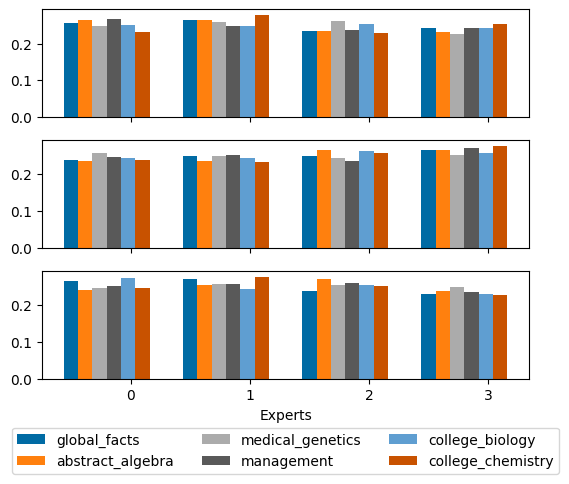}}
  \caption{Expert assignment results when evaluating on MMLU dataset using different pre-trained models on OpenWebText dataset. From left to right: (1) learned routing; (2) frozen routing; (3) random routing. }
 \label{fig: random-specialization}
\end{figure}

We visualize expert assignments when routing with our pre-trained model (Top-2 Layer-wise Sequence-level routing) on OpenWebtext data in Figure~\ref{fig: random-specialization}. Frozen routing and learned routing gave similar routing results, indicating weak expert specialization in topics. In the middle panel of Figure~\ref{fig: random-specialization}, we notice the co-existence of a common expert and weak specialized experts, aligning the design choice of shared experts in \cite{dai2024deepseekmoe}.

\begin{figure}[h!]
  \centering
   \subfigure{\includegraphics[width=0.31\textwidth]{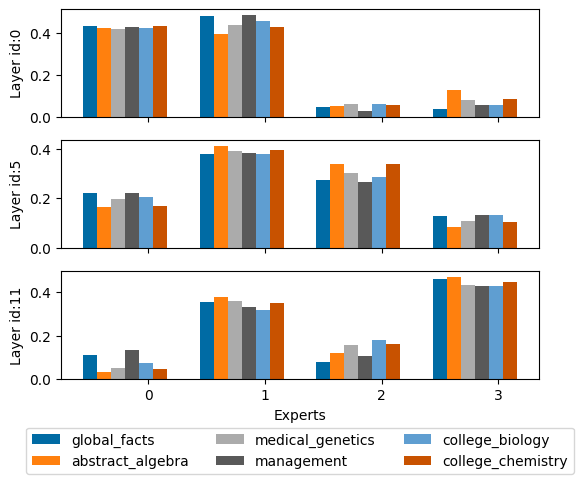}}
  \subfigure{\includegraphics[width=0.3\textwidth]{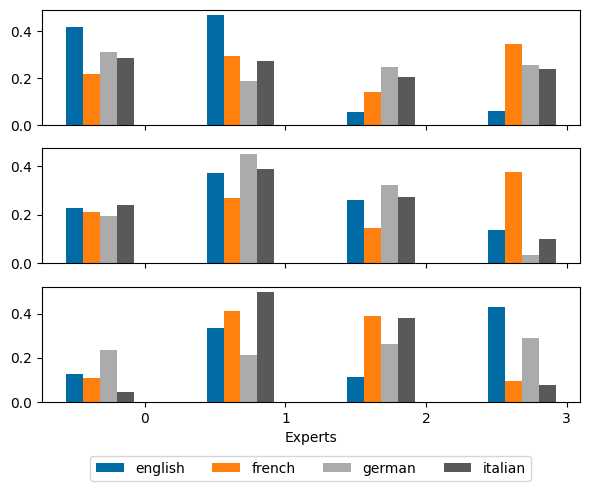}}
  \caption{Left: pre-trained on Multilingual Wikipedia dataset, evaluated on 6 categories of MMLU dataset; Right: pre-trained on Multilingual Wikipedia dataset, evaluated on 4 languages from XNLI dataset and X-Stance Dataset. Layer-wise Token-level Top2 routing is employed in pretraining.}
 \label{fig: token-level-routing-specialization}
\end{figure}

A visualization of expert assignments when routing with our pre-trained model (Top-2 Layer-wise Token-level routing) on Multilingual Wikipedia dataset is shown in Figure~\ref{fig: token-level-routing-specialization}. In contrast to weak topic specialization (Left panel of Figure~\ref{pic: Layer-wise expert assignment}) when routing at a sequence level, Token-level routing does not bring any expert specialization in topic domains, as suggested in the left plot of Figure~\ref{fig: token-level-routing-specialization}. The almost even expert assignment indicates the lack of expert specialization, similar to the observations in random routing illustrated in Figure \ref{fig: random-specialization}. The main difference arises from the balanced expert load in random routing, as opposed to the uneven distributions found in learned routing. However, we observe uneven assignments of various languages to different experts, likely stemming from differences in language tokenization.

\subsection{Layer wise expert activation}
\label{append-layerwise-expert-activation}
\begin{figure}
    \centering
\includegraphics[width=\textwidth]{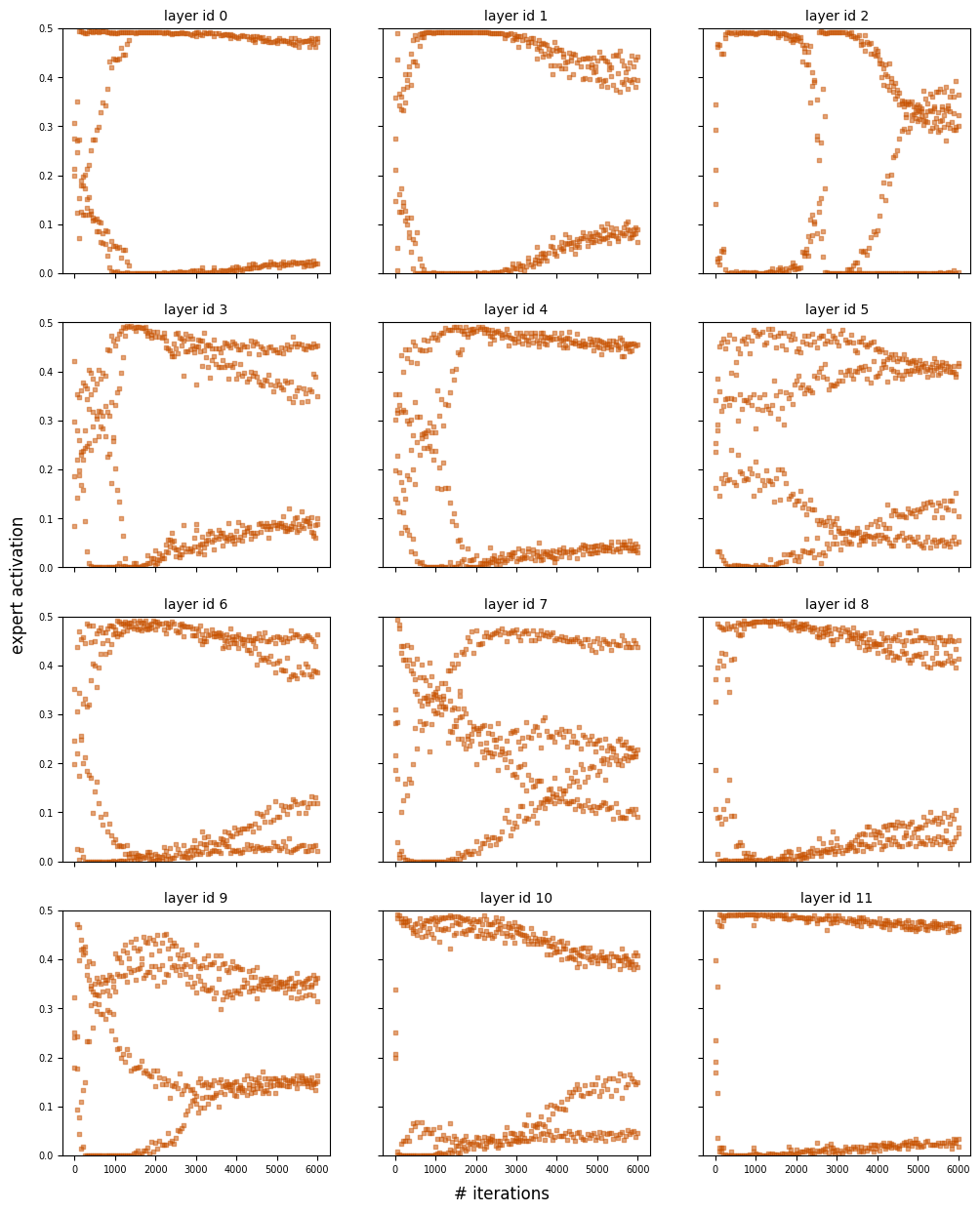}
    \caption{Expert activations from Layer-wise Sequence-level Top-2 routing when no load balancing loss is applied.}
    \label{fig:layer-wise-activations-no-lb}
\end{figure}

We visualize Layer-wise expert activation when routing at a sequence level without load balancing loss in Figure~\ref{fig:layer-wise-activations-no-lb}. When we refer to expert activation, we are describing the frequency at which each expert is activated. For each iteration (x-axis), there should be $N=4$ points denoting the corresponding activated frequency of each expert. We observe the following interesting aspects:
\begin{itemize}[leftmargin=*]
\item Experts exhibit the ability to recover from collapsing.

\item Early and late layers demonstrate a tendency to experience more expert collapse.
\end{itemize}

\begin{figure}
    \centering
\includegraphics[width=\textwidth]{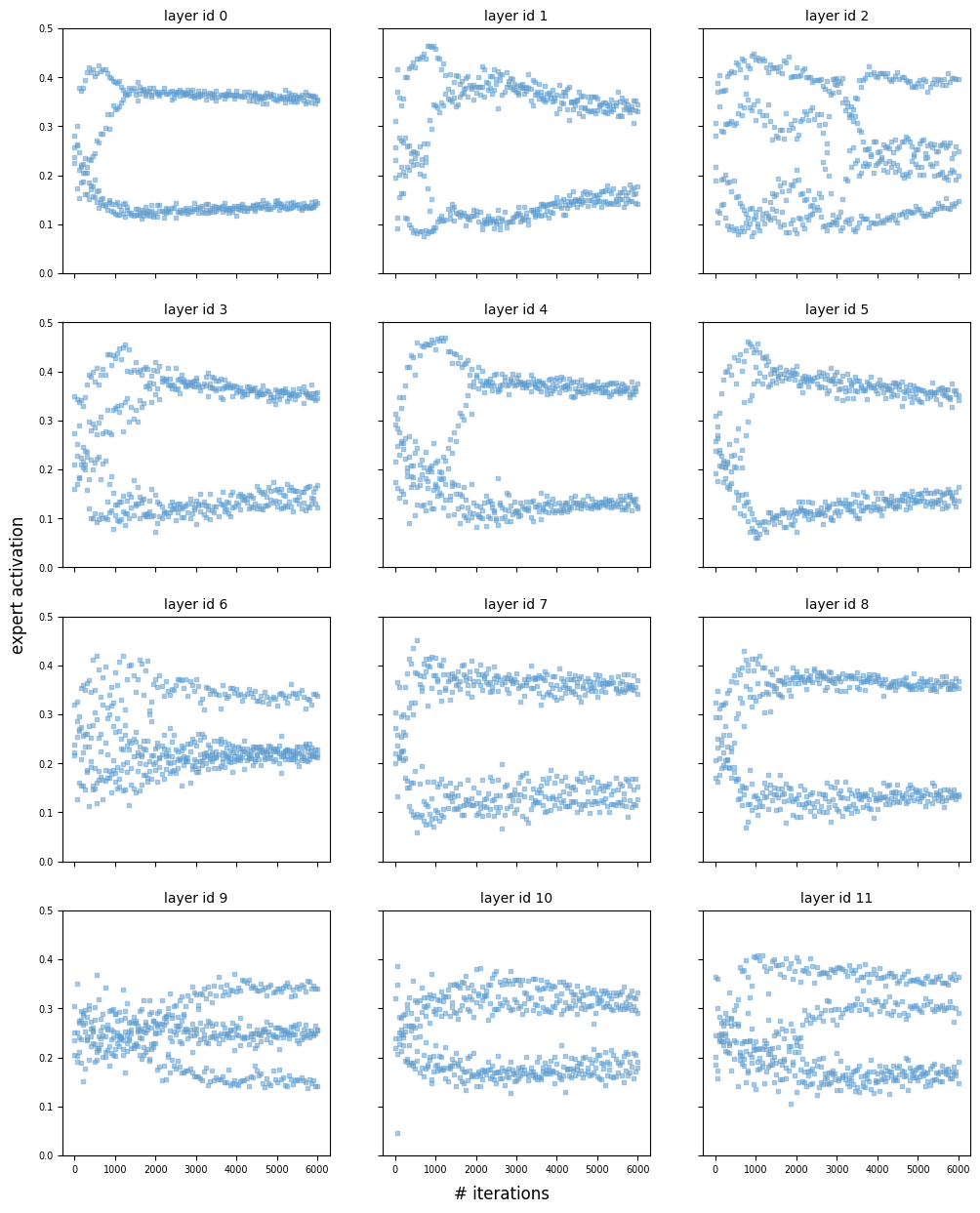}
    \caption{Expert activations from Layer-wise Sequence-level Top-2 routing when load balancing loss is applied.}
    \label{fig:layer-wise-activations-lb}
\end{figure}

\begin{figure}
    \centering
\includegraphics[width=\textwidth]{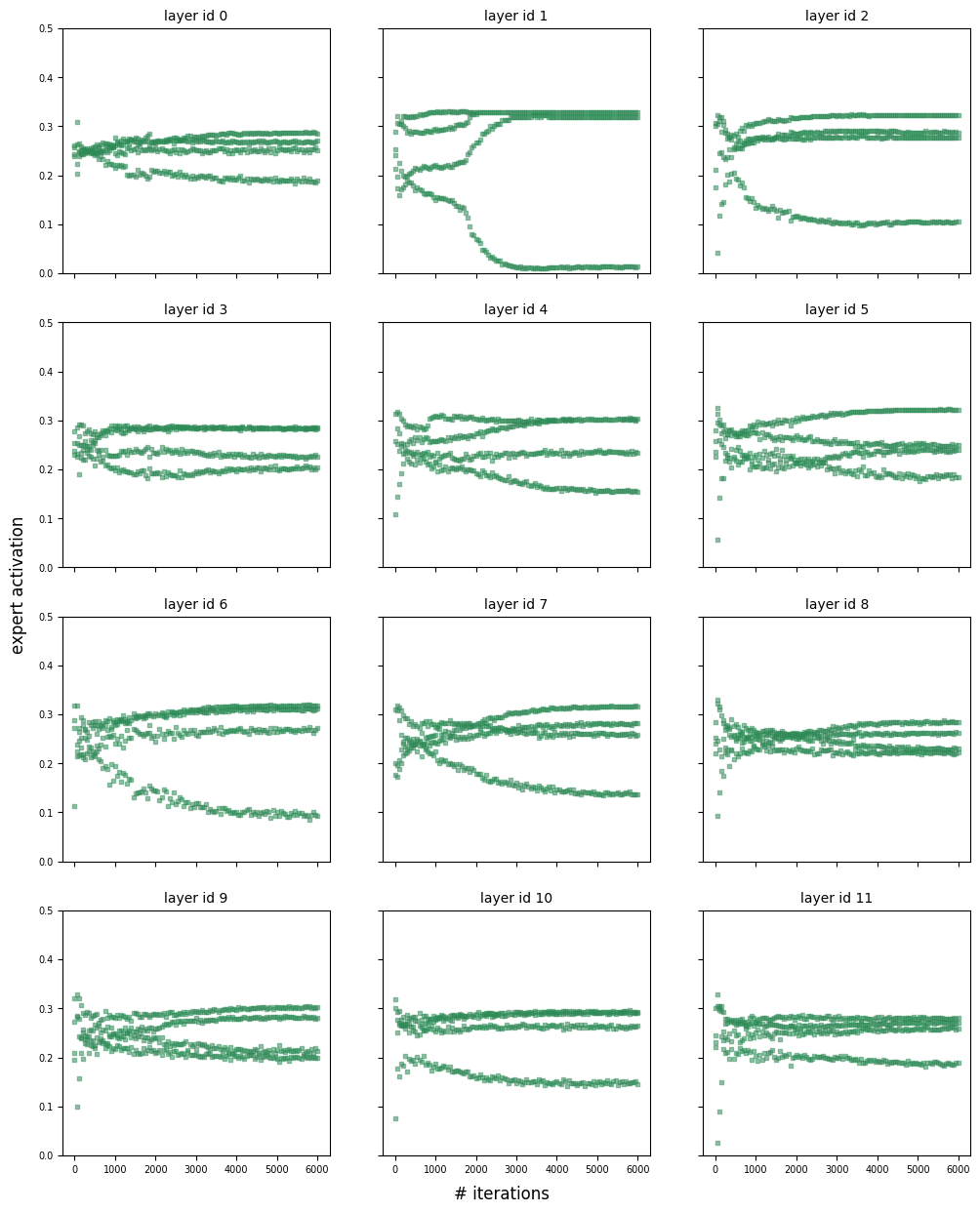}
    \caption{Expert activations from Layer-wise Sequence-level Top-3 routing when load balancing loss is applied.}
    \label{fig:layer-wise-activations-lb-top3}
\end{figure}

To provide readers with a clear visualization of layer-wise expert activations under the application of load balancing loss (with $\lambda$ = 0.01), we additionally plot the Top-2 and Top-3 routing results respectively in Figure~\ref{fig:layer-wise-activations-lb} and Figure~\ref{fig:layer-wise-activations-lb-top3}.

\end{document}

%% file: math_commands.tex
\usepackage{amsmath,amsfonts,bm}

\def\eqref#1{equation~\ref{#1}}

\def\1{\bm{1}}

\DeclareMathAlphabet{\mathsfit}{\encodingdefault}{\sfdefault}{m}{sl}
\SetMathAlphabet{\mathsfit}{bold}{\encodingdefault}{\sfdefault}{bx}{n}

\def\gT{{\mathcal{T}}}